\documentclass[conference]{IEEEtran}
\IEEEoverridecommandlockouts

\usepackage[colorlinks=true]{hyperref}
\hypersetup{
    colorlinks,
    citecolor=blue,
    linkcolor=blue
}

\usepackage{cite}
\usepackage[caption=false, font=footnotesize]{subfig}
\usepackage{amsmath,amssymb,amsfonts}
\usepackage{algorithmic}
\usepackage{graphicx}
\usepackage{textcomp}
\usepackage{xcolor}
\usepackage{color,soul}
\usepackage{colortbl}
\usepackage{tikz}
\usepackage{ctable}  
\usepackage{amsfonts, bm}
\usepackage{algorithm}
\usepackage{multirow}
\usepackage{mwe}

\newrobustcmd*{\myVtriangle}[2]{\tikz{\filldraw[draw=#1,fill=#2] (0cm,0.2cm) --
(0.2cm,0.2cm) -- (0.1cm,0cm) -- (0cm,0.2cm);}}

\newrobustcmd*{\mythickVtriangle}[2]{\tikz{\filldraw[line width=0.3mm,draw=#1,fill=#2] (0cm,0.2cm) --
(0.2cm,0.2cm) -- (0.1cm,0cm) -- (0cm,0.2cm);}}

\newrobustcmd*{\mythickErrorVtriangle}[2]{\tikz{\filldraw[line width=0.3mm,draw=#1,fill=#2] (-0.05cm,0.05cm) --
(0.05cm,0.05cm) -- (0cm,-0.05cm) -- (-0.05cm,0.05cm);  \draw[draw=#1] (0.0cm, -0.12cm) -- (0.0cm, 0.12cm) ; \draw[draw=#1] (-0.06cm, 0.12cm) -- (0.06cm, 0.12cm); \draw[draw=#1] (-0.06cm, -0.12cm) -- (0.06cm, -0.12cm)    }}

\newrobustcmd*{\mytriangle}[2]{\tikz{\filldraw[draw=#1,fill=#2] (0.0cm,0.0cm) --
(0.2cm,0cm) -- (0.1cm,0.2cm) -- (0cm,0cm);}}

\newrobustcmd*{\mysquare}[2]{\tikz{\draw[draw=#1,fill=#2] (0cm,0cm)
rectangle (0.2cm,0.2cm)}}

\newrobustcmd*{\mythicktriangle}[2]{\tikz{\filldraw[line width=0.3mm,draw=#1,fill=#2] (0.0cm,0cm) --
(0.2cm,0cm) -- (0.1cm,0.2cm) -- (0.0cm,0cm);}}

\newrobustcmd*{\mythicksquare}[2]{\tikz{\draw[line width=0.3mm,draw=#1,fill=#2] (0cm,0cm)
rectangle (0.2cm,0.2cm)}}

\newrobustcmd*{\mybarredtriangle}[2]{\tikz{\draw[draw=#1,fill=#2] (0,0) --
(0.2cm,0) -- (0.1cm,0.2cm) -- (0cm,0cm); \draw[draw=#1] (-0.1cm, 0.07cm) -- (0.3cm, 0.07cm)}}

\newrobustcmd*{\mythickbarredtriangle}[2]{\tikz{\draw[line width=0.3mm,draw=#1,fill=#2] (0,0) --
(0.2cm,0) -- (0.1cm,0.2cm) -- (0cm,0cm); \draw[draw=#1] (-0.1cm, 0.07cm) -- (0.3cm, 0.07cm)}}

\newrobustcmd*{\mybarredsquare}[2]{\tikz{\draw[draw=#1,fill=#2] (0,0)
rectangle (0.2cm,0.2cm); \draw[draw=#1] (-0.1cm, 0.1cm) -- (0.3cm, 0.1cm)}}

\newrobustcmd*{\mythickbarredsquare}[2]{\tikz{\draw[line width=0.3mm,draw=#1,fill=#2] (0,0)
rectangle (0.2cm,0.2cm); \draw[draw=#1] (-0.1cm, 0.1cm) -- (0.3cm, 0.1cm)}}

\newrobustcmd*{\mybarredcircle}[2]{\tikz{\draw[draw=#1,fill=#2] (0,0)
circle (0.1cm); \draw[draw=#1] (-0.2cm, 0.0cm) -- (0.2cm, 0.0cm)}}

\newrobustcmd*{\mythickbarredcircle}[2]{\tikz{\draw[line width=0.3mm,draw=#1,fill=#2] (0,0)
circle (0.1cm); \draw[draw=#1] (-0.2cm, 0.0cm) -- (0.2cm, 0.0cm)}}

\newrobustcmd*{\mythickErrorcircle}[2]{\tikz{\draw[line width=0.3mm,draw=#1,fill=#2] (0,0)
circle (0.06cm); \draw[draw=#1] (0.0cm, -0.12cm) -- (0.0cm, 0.12cm) ;   \draw[draw=#1] (-0.06cm, 0.12cm) -- (0.06cm, 0.12cm); \draw[draw=#1] (-0.06cm, -0.12cm) -- (0.06cm, -0.12cm)    }}

\newrobustcmd*{\mydashedline}[1]{\tikz{\draw[draw=#1] (-0.2cm, 0.2cm) -- (-0.1cm, 0.2cm); \draw[draw=#1] (-0.0cm, 0.2cm) -- (0.1cm, 0.2cm)}}

\newrobustcmd*{\mythickcross}[1]{\tikz{\draw[line width=0.3mm,draw=#1] (0,0) --
(0.2cm,0); \draw[line width=0.3mm,draw=#1] (0.1cm,-0.1cm) -- (0.1cm,0.1cm);}}

\newrobustcmd*{\mybarredcross}[1]{\tikz{\draw[line width=0.3mm,draw=#1] (0,0) --
(0.2cm,0); \draw[line width=0.3mm,draw=#1] (0.1cm,-0.1cm) -- (0.1cm,0.1cm); \draw[draw=#1] (-0.1cm,0) -- (0.3cm,0);}}

\newrobustcmd*{\myline}[1]{\tikz{\draw[draw=#1] (-0.15cm, 0.1cm) -- (0.15cm, 0.1cm);\draw[line width=0.3mm,draw=#1] (-0.0cm, 0.0cm);}}

\newrobustcmd*{\mythickline}[1]{\tikz{\draw[line width=0.3mm,draw=#1] (-0.15cm, 0.1cm) -- (0.15cm, 0.1cm);\draw[line width=0.3mm,draw=#1] (-0.0cm, 0.0cm);}}

\newrobustcmd*{\mythickdashedline}[1]{\tikz{\draw[line width=0.3mm,draw=#1] (-0.2, 0.1cm) -- (-0.1cm, 0.1cm); \draw[line width=0.3mm,draw=#1] (-0.0cm, 0.1cm) -- (0.1cm, 0.1cm); \draw[line width=0.3mm,draw=#1] (-0.0cm, 0.0cm);}}

\newrobustcmd*{\mythickdasheddottedline}[1]{\tikz{\draw[line width=0.3mm,draw=#1] (-0.22, 0.1cm) -- (-0.13cm, 0.1cm); \draw[line width=0.3mm,draw=#1] (-0.085cm, 0.1cm) -- (-0.055cm, 0.1cm); \draw[line width=0.3mm,draw=#1] (-0.01cm, 0.1cm) -- (0.08cm, 0.1cm); \draw[line width=0.3mm,draw=#1] (-0.0cm, 0.0cm);}}

\newrobustcmd*{\mycircle}[2]{\tikz{\draw[draw=#1,fill=#2] (0,0)
circle (0.1cm);}}

\newrobustcmd*{\mythickcircle}[2]{\tikz{\draw[line width=0.3mm,draw=#1,fill=#2] (0,0)
circle (0.1cm);}}

\newrobustcmd*{\mydot}[1]{\tikz{\draw[line width=0.3mm,draw=#1] (0,0)
circle (0.025cm);}}

\definecolor{darkspringgreen}{rgb}{0.09, 0.45, 0.27}

\begin{document}

\bstctlcite{IEEEexample:BSTcontrol}

\title{Discovery of False Data Injection Schemes on Frequency Controllers with Reinforcement Learning \\

\thanks{$^{1}$National Renewable Energy Laboratory (NREL), Golden, CO, USA.}%
\thanks{$^{2}$College of Engineering \& Computer Science, Syracuse University, Syracuse, NY, USA. Work completed during Romesh's internship at NREL.}%
\thanks{This work was authored by the National Renewable Energy Laboratory (NREL), operated by Alliance for Sustainable Energy, LLC, for the U.S. Department of Energy (DOE) under Contract No. DE-AC36-08GO28308. This work was supported by the Laboratory Directed Research and Development (LDRD) Program at NREL. The views expressed in the article do not necessarily represent the views of the DOE or the U.S. Government. The U.S. Government retains and the publisher, by accepting the article for publication, acknowledges that the U.S. Government retains a nonexclusive, paid-up, irrevocable, worldwide license to publish or reproduce the published form of this work, or allow others to do so, for U.S. Government purposes.}
\thanks{This research was performed using computational resources sponsored by the Department of Energy's Office of Energy Efficiency and Renewable Energy and located at the National Renewable Energy Laboratory.}
}

\author{\IEEEauthorblockN{Romesh Prasad$^{1,2}$, Malik Hassanaly$^1$, Xiangyu Zhang$^1$ and Abhijeet Sahu$^1$}}

\maketitle

\begin{abstract}
While inverter-based distributed energy resources (DERs) play a crucial role in integrating renewable energy into the power system, they concurrently diminish the grid's system inertia, elevating the risk of frequency instabilities. Furthermore, smart inverters, interfaced via communication networks, pose a potential vulnerability to cyber threats if not diligently managed. To proactively fortify the power grid against sophisticated cyber attacks, we propose to employ reinforcement learning (RL) to identify potential threats and system vulnerabilities. This study concentrates on analyzing adversarial strategies for false data injection, specifically targeting smart inverters involved in primary frequency control. Our findings demonstrate that an RL agent can adeptly discern optimal false data injection methods to manipulate inverter settings, potentially causing catastrophic consequences.   
\end{abstract}


\section{Introduction}

The increasing integration of distributed energy resources (DERs) helps introduce more clean energy and flexibility to the power system and plays a pivotal role in decarbonizing the future energy systems \cite{horowitz2019overview}. Under this trend, future power systems are transitioning from traditional all-synchronous generator (SG) systems to a mixture of SGs and a large number of inverter-based resources (IBRs). With a higher penetration level of IBRs, the power grid inertia will reduce, leading to an elevated concern of frequency stability. Interconnected smart inverters are a promising strategy to address the issue of inertia reduction, at the expense of emerging cybersecurity concerns. If interconnected smart inverters are compromised, they could serve as an interface for initiating cyber-adversarial events that may lead to catastrophic results.

There is an increasing interest in investigating how adversarial events can affect power systems by infiltrating and controlling smart inverters. For example, how smart inverters' settings can be altered by false data injection (FDI) under the operation mode of Volt-VAR, Volt-Watt, and constant power factor is studied in \cite{olowu2020investigation}. Detection and mitigation strategies for cyber-attacks such as FDI and denial-of-service (DOS) targeting the automatic generation control (AGC) system were developed in \cite{roy2019detection}. Tuyen et al.~\cite{tuyen2022comprehensive} presented a review of the cybersecurity risks for inverter-based power systems, including typical attacks, defense mechanisms, and detection and mitigation measures. Similar reviews can also be found in \cite{li2022cybersecurity, sahoo2019cyber}. Additionally, cyber-physical system (CPS) testbeds are developed to further facilitate research in this area: in ~\cite{liu2015analyzing}, DoS attacks are executed in a testbed equipped with a real-time digital simulator (RTDS) for power, the network simulator-3 (NS-3) for communications, and devices like phasor measurement units and phasor data concentrators. This setup aims to investigate the vulnerability of voltage stability monitoring and control in a transmission system to these attacks. The work presented in \cite{zhang_2021} demonstrates that as the communication delay for a photovoltaic system exceeds the maximum tolerable level, it leads to a severe power imbalance and triggers an over-voltage event.

The escalating complexity of problem formulations and the intertwining of both physical and cyber systems have motivated researchers to seek solutions through machine learning-based approaches. The remarkable performance of reinforcement learning (RL) in sequential decision-making across various applications has garnered attention for its potential to address challenges within cyber-physical systems. For example, RL was used for automated incident handling against network-based attack~\cite{ossenbuhl2015towards}. Galowicz et al.~\cite{gawlowicz2019ns}, trained a communication network controller to take action based on signals such as signal-to-noise ratio thresholds. 

For power system frequency control, RL has been utilized for secondary control such as single-area or multi-area AGC. For instance, a twin delayed deep deterministic policy gradient (DDPG) method is used to improve the performance of a single-area AGC in~\cite{single_area_agc}. A multiagent DDPG technique is employed for multi-area load frequency control~\cite{multi_area_agc}. Overall, RL strategies have mostly focused on improving load recovery and control of the power grid.

In this work, we propose using an RL strategy from an adversary's perspective to discover hitherto unseen vulnerabilities. The vulnerability targetted is the risk that FDI could sufficiently compromise the frequency controller to induce frequency instabilities. We investigate the use of RL to learn optimal FDI adversarial strategies against the default droop control in the smart inverter. Equipped with the knowledge of possible FDI strategies, power grid operators and planners could preemptively implement countermeasures. Specifically, our contributions include:

\begin{enumerate}
    \item Proposing an RL-based power system vulnerability identification method that can be generalized to other problems.
    \item Demonstrated in the inverter-based power system case study that the proposed approach can adeptly discern optimal FDI actions that lead to catastrophic consequences.
    \item Identifying hurdles in ensuring reliable training of the adversary training agents. 
\end{enumerate}

The rest of the paper is organized as follows: Section~\ref{sec:prob} describes the problem formulation that helps design the adversarial strategy against power system frequency control; Section~\ref{sec:vul-disc} proposes an RL-based mechanism for discovering vulnerability; Section~\ref{sec:results} presents key results on a frequency control case study, and analyzes the policy learned by the RL agent.

\section{Problem Formulation}
\label{sec:prob}

This section presents the specific power system frequency control problem as well as the formulation of the interested FDI attacks.

\subsection{Power System Frequency Dynamics}

The system frequency dynamics can be modeled by the swing equation \cite{kundur2022power}:
\begin{subequations}  \label{eq-swing-equation}
\begin{align}
    \dot{\theta_i} &= \omega_i,\\
    M_i\dot{\omega_i} &= p_{i}-p_{e, i}-D_i \omega_i-p^   {\text{IBR}}_i
\end{align}
\end{subequations}
in which $i\in\mathcal{N}=\{1,..., n\}$ is the bus index, $\theta_i$ and $\omega_i$ are the voltage phase angle and frequency deviation of bus $i$ respectively. $M_i$ and $D_i$ are the inertia and damping coefficients. The net power injection is denoted by $p_i$. By applying the direct current (DC) approximation in transmission system modeling, there is electric power $p_{e, i}=\sum_{j\in(\mathcal{N}\setminus\{i\})}B_{ij} \text{sin} (\theta_i - \theta_j)$ and $B_{ij}$ is the susceptance of line between bus $i$ and $j$. $p^{\text{IBR}}_i$ represents the power output from the inverter-based resource on bus $i$. 

\begin{figure}
    \centering
    \includegraphics[width=0.99\linewidth]{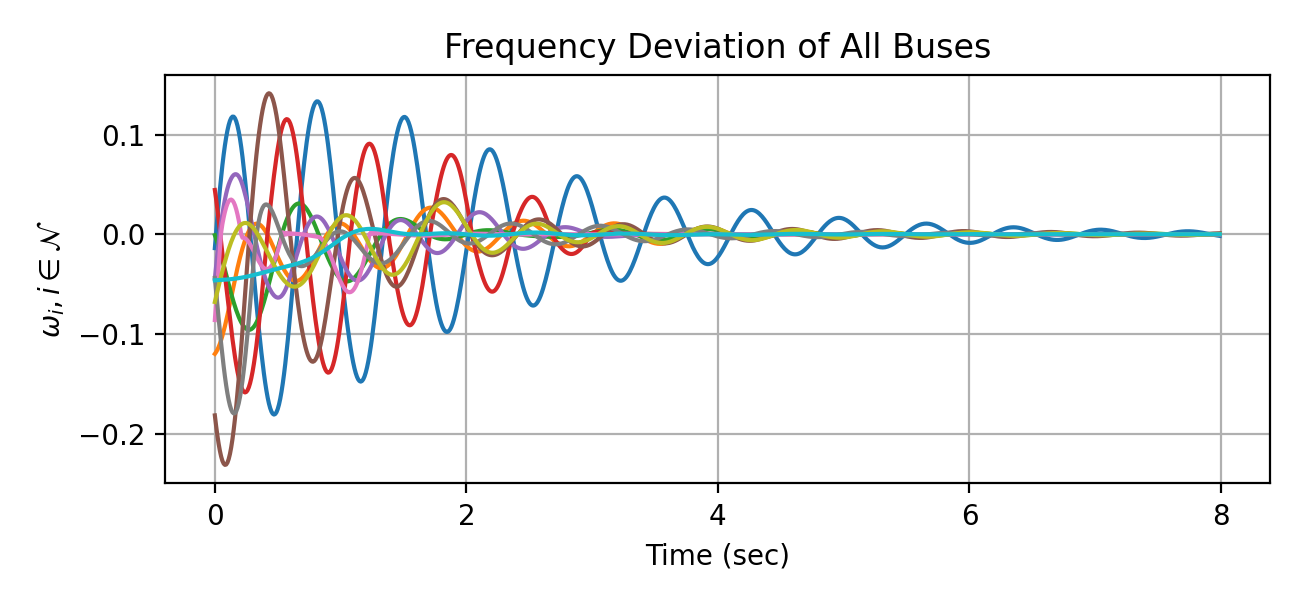}
    \caption{IBRs following properly designed droop control to stabilize frequency.}
    \label{fig-normal-droop}
    \vspace{-1em}
\end{figure}

To maintain the system's frequency stability, the fast responding IBRs on all $n$ buses participate in the system primary frequency control (PFC) following their designed droop rule, i.e., $p^{\text{IBR}}_i=k_i \omega_i$. For simplicity, it is assumed that the IBRs have enough headroom to respond to both high and low-frequency events. Droop coefficients $k_i$ are properly designed so that under normal disturbances, the system frequency can be stabilized. The system considered throughout this work is the Kron-reduced IEEE New England transmission system with $n=10$ buses on which both traditional synchronous machines and IBRs exist. The numerical integration of Eq.~\ref{eq-swing-equation} is conducted with an explicit Euler integration with timestep $\Delta t = 0.01s$ similar to \cite{cui2022reinforcement}. Fig.~\ref{fig-normal-droop} demonstrates the effect of the frequency control starting from a random off-equilibrium initial condition.

\subsection{Adversarial Strategy Design}

Through the cyber-connected smart inverters, an attacker could gain access to these key components of the power system and alter their internal logic to cause catastrophic disturbances. In this study, it is assumed that an attacker has already infiltrated the system allowing to tamper with the control logic of the droop controller by modifying droop coefficients   (i.e. $p^{\text{IBR}}_i=k^{'}_i \omega_i$, where $k^{'}_i$ are the modified droop coefficients). Other types of data tampering could also be envisioned and are left for future work \cite{nguyen2020electric}. The objective of the attacker is to maximize the disturbance created, specifically, given by: 

\begin{equation} \label{eq-adervary-obj}
\begin{aligned}
& \underset{k_{i, t}^{'}, i\in\mathcal{N}, t\in\mathcal{T}}{\operatorname{maximize}} 
& & \sum_{i\in\mathcal{N}} \sum_{t\in\mathcal{T}} \delta\omega_{i, t} \\
& \text{subject to} && \eqref{eq-swing-equation}, \\
&&& p^{\text{IBR}}_{i, t}=k^{'}_{i, t} \omega_{i, t}, \forall t \\
&&&  ||k^{'}_{i, t} - k_{i,t}||_0 \leq 1, \forall t
\end{aligned}
\end{equation}
where $\delta\omega_{i, t}=|\omega_{i,t}| - |\omega_{i,t}^{\text{base}}|$ represents the difference between the frequency deviation after the FDI action $\omega_{i,t}$ and the baseline without any attack $\omega_{i,t}^{\text{base}}$ after a major frequency disturbance event. The design of this objective function is based on the intuition of maximizing additional disturbance by injecting false data, thereby exacerbating the consequences in the event of a system disturbance. To keep the FDI stealthy, the following constraint is placed on the type of FDI that can be conducted.  
\textit{\textbf{Assumption}}: The adversary can only target one IBR at each control step, i.e., for any $t\in\mathcal{T}$, there can be at most one $i$ with $k_i^{'}\neq k_i$. The stealth requirements are captured by the third constraint in Eq.~\ref{eq-adervary-obj}.

In future works, additional requirements can be added to make the formulation more realistic, including that the tampering on any single IBR should not last more than $t_{\text{detect}}$ seconds cumulatively over $\mathcal{T}$ to avoid being detected. 

\section{Vulnerability Discovery via RL}
\label{sec:vul-disc}

\begin{figure}
    \centering
    \includegraphics[width=0.99\linewidth]{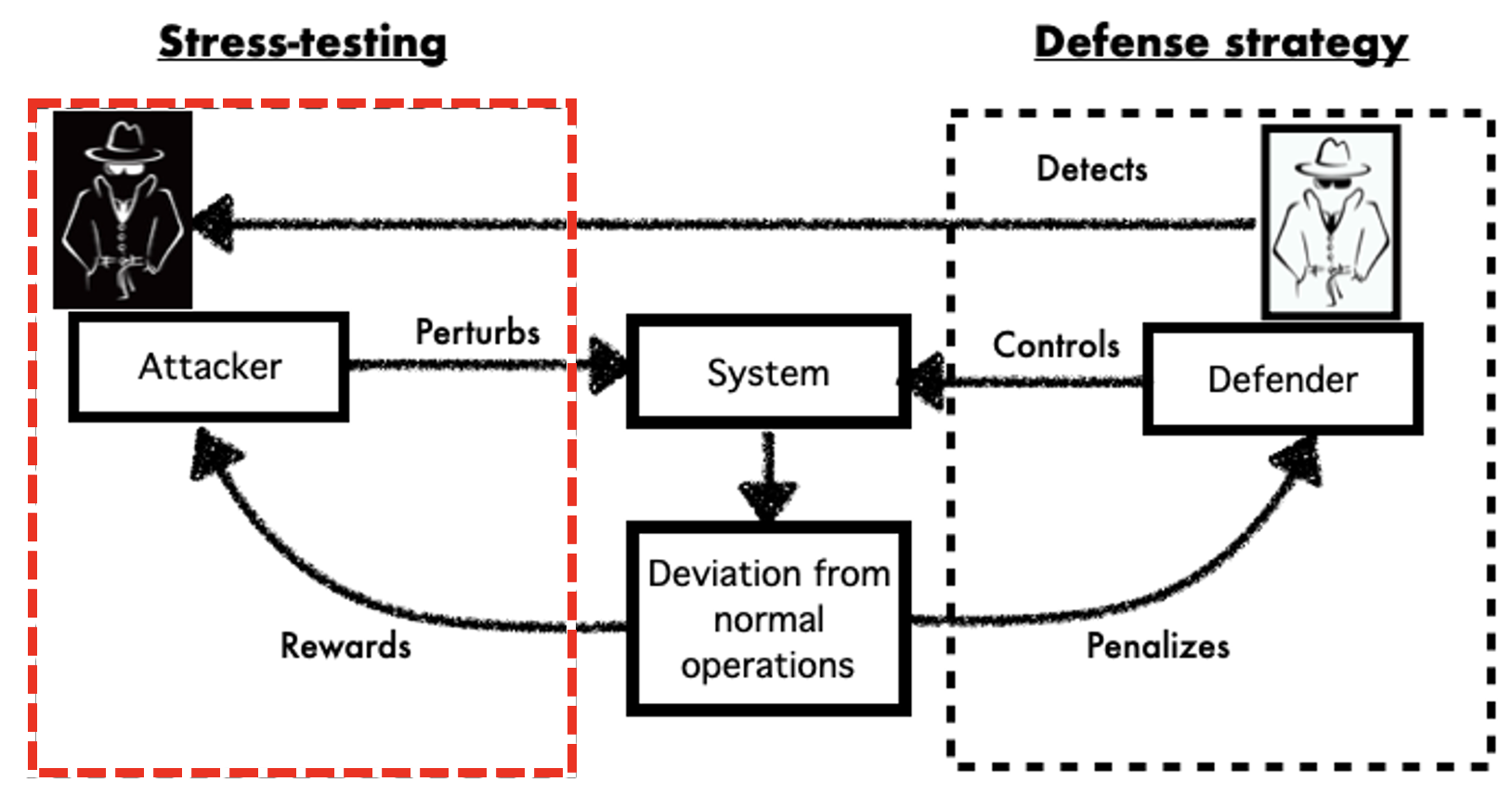}
    \caption{An adversarial design process for vulnerability discovery and defense strategies development for adversarial events in CPS. The red box indicates the focus of this paper.}
    \label{fig-overview}
    \vspace{-1em}
\end{figure}

To discover possible system vulnerability to FDI, we adopt the perspective of an adversary by train an RL agent to induce frequency instabilities through droop coefficient tampering. This vulnerability discovery process is depicted in the dashed red box in Fig.~\ref{fig-overview}, as the focus of this paper; once vulnerabilities are identified, corresponding defense strategies could be devised to prevent elaborate FDI strategies. The defense mechanism and this overall adversarial designing process will be formally studied in our future works.

\subsection{Markov Decision Process Formulation}
A Markov decision process (MDP) needs to be formulated \cite[Chapter~3]{sutton_barto_2018}, by defining the observation, action, and reward, in order to use RL to solve the aforementioned sequential decision-making problem. \color{black}

\textbf{Observation} $\mathbf{s}_t$ collects all the information the adversary uses for decision-making. Here, for simplicity, it is assumed that the adversary has already gained full observability of the system. So, there is $\mathbf{s}_t=[\bm{\omega}_t^{\mathcal{N}}, \bm{\theta}_t^{\mathcal{N}}]^\top$, with voltage phase angle and frequency deviation of all buses at step $t$. 

\textbf{Action} $\mathbf{a}_t=[i, k_i^{'}]^{\top}$ is multi-categorical, it chooses which IBR to disturb ($i \in \mathcal{N}$) and how to tamper the droop coefficient ($k\in\kappa_i$).

\textbf{Reward} $r_t$ is defined corresponding to the objective function in \eqref{eq-adervary-obj}, which reflects the frequency deviation relative to that of the original trajectory at step $t$: $r_t=\mathbf{1}^{\top} (\bm{\omega}_t^{\mathcal{N}} - \bm{\omega}_t^{\mathcal{N}, \text{original}})$.

Based on the MDP defined above, an environment can be built to train an RL agent to maximize the cumulative reward $\mathcal{R} = \sum_{t\in\mathcal{T}} r_t$, which quantifies the intensity of the frequency instability induced. 

\subsection{Reinforcement Learning Algorithm}
Due to its promising performances in other contexts \cite{yu2022surprising}, the proximal policy optimization (PPO) algorithm is used to train the adversarial policy \cite{schulman2017proximal}. As a separate analysis (not shown here for the sake of brevity), an asynchronous actor-critic method \cite{mnih2016asynchronous} was also employed and led to lower cumulative reward than PPO. 
Within the PPO algorithm, the training maximizes the following surrogate objective:

\begin{equation}
\begin{aligned}
    \mathcal{L}(\phi)&=\mathbb{E}[\text{min}(y_t(\phi)A_t, \text{clip}(y_t(\phi), 1-\epsilon, 1+\epsilon)A_t) \\
    &-c_V\mathcal{L}_V(\phi) + c_H H_{\phi}(\mathbf{s}_t)]
\end{aligned}
\end{equation}
where $y_t(\phi)=\frac{\pi_{\phi}(\mathbf{a}_t|\mathbf{s}_t)}{\pi_{\phi_{\text{old}}}(\mathbf{a}_t|\mathbf{s}_t)}$ is the probability ratio, which we would like to increase if the advantage $A_t$ is positive and decrease if $A_t$ is negative. Clipping it by $1\pm\epsilon$ encourages policy improvement within a trust region. $\mathcal{L}_V(\phi)$ is the value function loss, indicating how accurately the critic can estimate the expected reward. $H_{\phi}$ is an entropy term that encourages the action diversity of the policy. The hyperparameters $c_V$ and $c_H$ control the relative importance of value function loss and entropy when compared with the policy loss. The policy is encoded with a neural net with 2 hidden layers of 64 neurons each. Batches of 64 environment steps are used for training. The entropy loss coefficient is chosen to be $c_H=0.001$, the value loss coefficient is $c_V = 0.5$ and the learning rate is set constant equal to $3\times 10^{-4}$.

\section{Case Study}
\label{sec:results}

\subsection{Experimental Setup}

The case study is conducted using the 10-bus Kron reduced IEEE New England system, with the implementation of the system frequency dynamics and default droop settings adopted from Ref.~\cite{cui2022reinforcement}. The software implementation of the RL algorithm relies on Stable-Baselines3 \cite{stable-baselines3}, and the training is done on the NREL HPC system. The environment defined by Eq.~\ref{eq-swing-equation} is simulated for 500 steps (equivalent to 5s of total simulation time) and the objective of the agent is to maximize the cumulative reward over these 500 steps. The initial conditions for the phase and frequency of the buses are uniformly randomly sampled from a multivariate uniform distribution $\mathcal{U}(-0.03, 0.03)$ superimposed with the equilibrium state of the system. The same initial condition is used for all realizations of the environment in order to guarantee a deterministic map between action and cumulative reward.

At every step, the agent chooses any one of the 10 generators and chooses to replace its original droop controller coefficient $k_i$ with $k'_i$ for 1 step. For simplicity, $k'_i \in \{ -1, 0, 1 \}$ independently of the original value $k_i$.

\subsection{Validation with a simplified action space}
\label{sec:timeInv}
\begin{figure}
    \centering
    \includegraphics[width=0.99\linewidth]{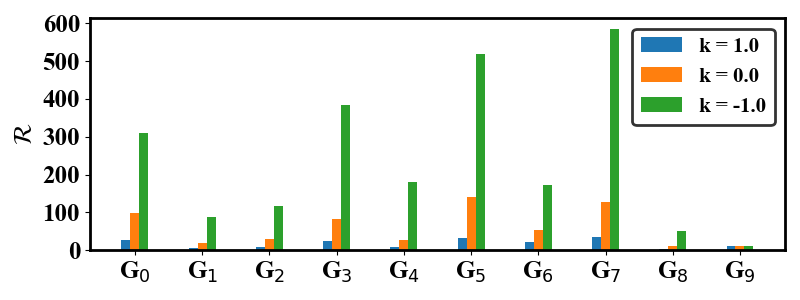}
    \caption{Cumulative reward obtained with a time-invariant policy for all 30 possible actions. $G_i$ denotes generator $i$.}
    \label{fig:manualFDI}
    \vspace{-1em}
\end{figure}

Albeit multidiscrete, the number of possible action is $\exp(\underbrace{500}_{\rm Num. steps} \log{\underbrace{30}_{\rm Num. actions}})$. The large action space prevents a brute force optimization strategy that would otherwise help validate the trained agent. Instead, a simplified action space is considered to derive a baseline performance evaluation for the policy obtained. The simplified action space chosen is a time-invariant action space, where the agent chooses one action type (one generator and one replacement value of the droop coefficient) throughout the 500 steps. Therefore, only 30 possible actions exist and can be easily enumerated to identify an optimal policy. The cumulative reward of the 30 different action types is shown in Fig.~\ref{fig:manualFDI}

Overall, it is clear that replacing the droop coefficient with a negative number results in the highest rewards. This is an expected result given that a negative droop coefficient would enhance frequency deviations rather than mitigate them. In addition, the cumulative reward strongly depends on the bus chosen for the FDI. Perturbing the bus 7 with an $k'_i=-1$ results in the highest cumulative reward $\mathcal{R} = 584$. By contrast, perturbations applied to $G_9$ have almost no effect on the cumulative reward. The optimal time-invariant policy and reward are used hereafter to evaluate the time-varying policy learned.

\subsection{RL training on the original action space}
\label{sec:origSpace}
\begin{figure}
    \centering
    \includegraphics[width=0.99\linewidth]{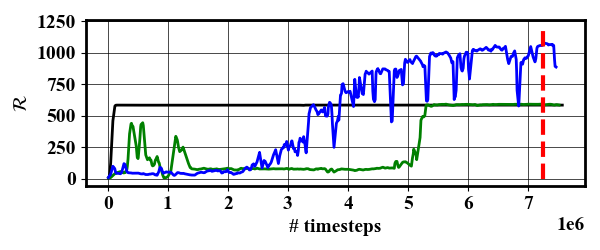}
    \includegraphics[width=0.99\linewidth]{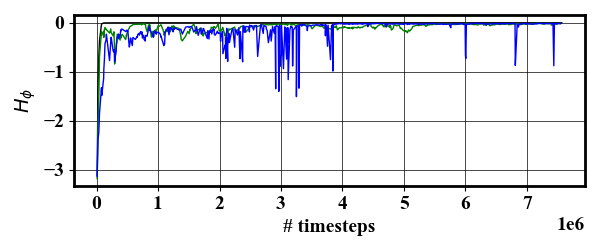}
    \caption{Top: cumulative reward history as a function of the number of environment steps simulated for 3 training runs (P1  \mythickline{black} matches the time-invariant reward, P2 \mythickline{darkspringgreen} slightly exceeded it, and P3 \mythickline{blue} reached the highest final reward) super-imposed time instance at which actions are recorded (\mythickdashedline{red}). Bottom: entropy loss history as a function of the number of environment steps simulated for the 3 training runs (\mythickline{black} for P1, \mythickline{darkspringgreen} for P2 and P3 \mythickline{blue}).}
    \label{fig:history}
    \vspace{-1em}
\end{figure}

The original action space has high dimensions as discussed in Sec.~\ref{sec:timeInv} which would make classical optimization strategies difficult to deploy and evaluate. In addition, any data-based method would inevitably require constructing a mapping between the action and the reward using a large dataset given the high dimension of the action space. The problem adopted is, therefore, naturally conducive to RL.
Three separate RL training runs are described to demonstrate the variability of the learned policy with respect to the initial neural net weights. The training history of the three runs is shown in Fig.~\ref{fig:history}. Significant variability is observed across the runs which suggests that local minima in the loss landscape might be difficult to escape. Out of the three runs shown hereafter, one matched the time-invariant cumulative reward (P1), and two exceeded it (P2 and P3) which suggests that the RL strategy is effective for the case considered. Notably, P3 achieved a time-invariant reward almost twice as high as the time-invariant baseline. The large oscillation of the highest-performing training run, also suggests that the map between action and reward might be particularly rough. Interestingly, the cumulative reward was initially one of the smallest. The entropy loss history suggests that this training realization explored the action space more than the other ones which eventually resulted in a higher $\mathcal{R}$.

\begin{figure}
    \centering
    \includegraphics[width=0.49\linewidth]{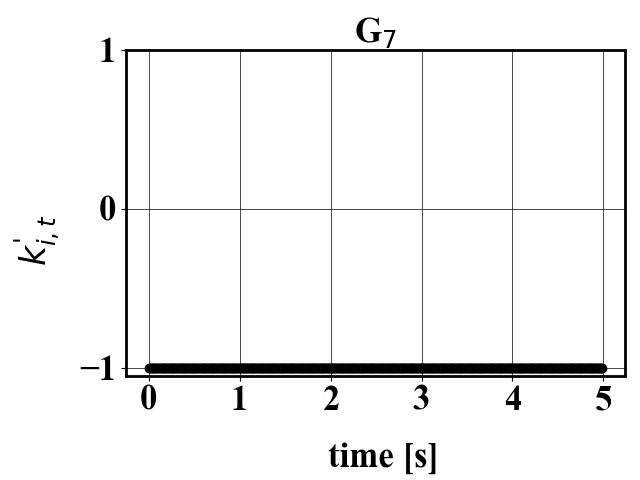}
    \includegraphics[width=0.49\linewidth]{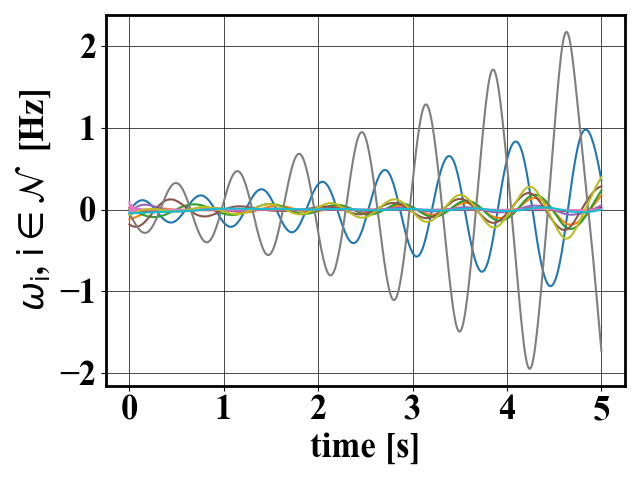}
    \includegraphics[width=0.49\linewidth]{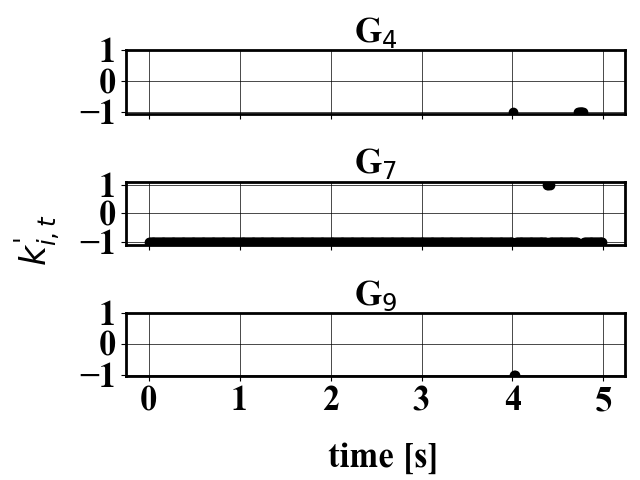}
    \includegraphics[width=0.49\linewidth]{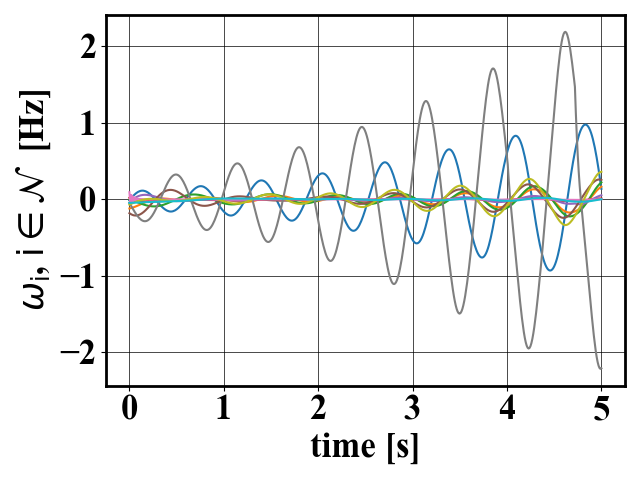}
    \includegraphics[width=0.49\linewidth]{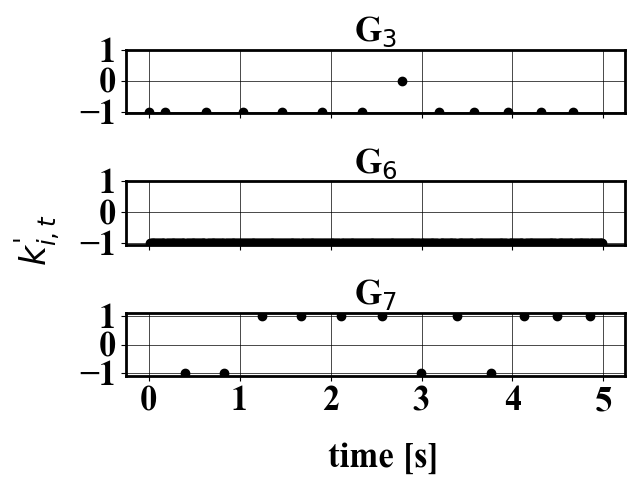}
    \includegraphics[width=0.49\linewidth]{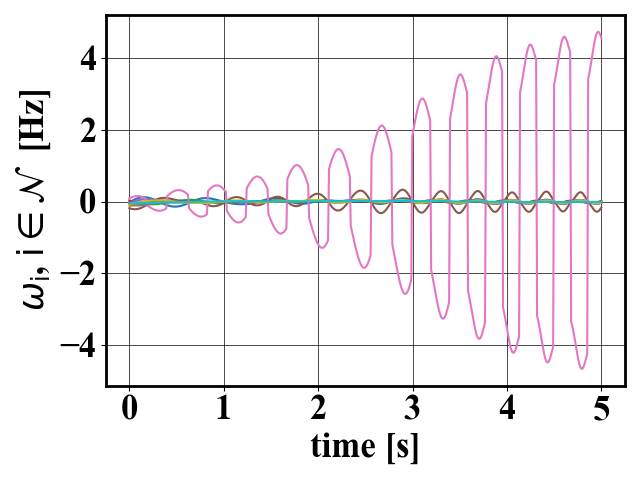}
    \caption{Learned policy (left) and system response (right) after $7.2 \times 10^6$ steps. Top: results for P1 (reward of 584). Middle: results for P2 (reward of 590). Bottom: results for P3 (reward of 1085). Only the generators activated by the policy are shown. The system response frequencies $\omega_i$ are shown for each generator.}
    \label{fig:policy}
    \vspace{-1em}
\end{figure}

The policies learned by the agents are extracted after $7.2 \times 10^6$ steps (dashed red line in Fig.~\ref{fig:history}), and shown in Fig.~\ref{fig:policy}. P1 achieved a reward of $584$, near that of the optimal time invariant policy (Sec.~\ref{sec:timeInv}). The P1 agent policy is shown in the top left of Fig~\ref{fig:policy}, along with the frequency response obtained (top right). Unsurprisingly, P1, has rediscovered the optimal time-invariant policy that was obtained in Sec.~\ref{sec:timeInv}, i.e. by constantly perturbing generator 7. The middle row of Fig.~\ref{fig:policy} shows the results of P2 that achieves a slightly higher reward (590). The P2 policy (middle left of Fig.~\ref{fig:policy}) is similar to the optimal time-invariant policy but also exercises intermittent perturbations on two other generators. Interestingly, the agent decides to perturb $G_9$ which has little effect in the time-invariant case study. This highlights the ability of the agent to discover counter-intuitive policies. In addition, the agent only switched between 3 of the 9 generators. This information would be valuable when planning cybersecurity measures tailored to each generator. The policy discovered exemplifies that the time-varying environment allows for an elaborate action scheduling that can induce higher perturbations than the time-invariant policy. In addition, the time-varying action can make the FDI detection more involved. The frequency response obtained with P2 (middle right of Fig.~\ref{fig:policy}) is similar to that of the time-invariant policy but manages to induce the onset of frequency instabilities earlier, thereby achieving a higher reward. Finally, the bottom row of Fig.~\ref{fig:policy} shows the P3 policy (bottom left) achieves higher frequency oscillations and of higher magnitude (bottom right), with a reward of 1085 by discovering a non-intuitive action. Instead of mostly perturbing $G_7$, the agent primarily perturbs $G_6$ and intermittently switches to $G_7$ and $G_3$ (Fig.~\ref{fig:policy} bottom left). Interestingly, the action is not always going against the droop controller (positive $k'_{i,t}$, which would be even more difficult to detect). Looking at the system response, it appears that the frequency at which the agent switches between generators lines up with the frequency deviation oscillation. However, the frequency deviation also appears to experience sudden jumps, which suggests that the agent also exploits numerical instabilities to achieve higher rewards. In the present case, the base droop controller coefficient was the highest for $G_6$ which makes the response of the unperturbed droop controller for $G_6$ large, if $|\omega_6|$ is large. The agent exploits this features by increasing $|\omega_6|$ via perturbation to $G_6$, before intermittently returning to the original droop coefficient for $G_6$, and perturbing $k_{7,t}$ instead. Finally, comparing the reward achieved to the time-invariant reward when perturbing only $G_6$ ($\mathcal{R} = 174$ as shown in Fig.~\ref{fig:manualFDI}), it is clear that the map between action and cumulative reward is indeed highly corrugated. Despite being able to choose actions that switch between generators with high frequency, the RL agent consistently chose to perturb the same generator for several steps. In practice, this type of action could lead to easier detection.

\section{Conclusion}

In this work, a reinforcement learning approach to discover hitherto unseen FDI schemes that would perturb power systems has been demonstrated on a model system. All realizations of the RL agent were able to match or exceed the optimal cumulative reward of a simpler action space. The RL agent consistently identified a subset of the buses that needed to be perturbed which could inform how to prioritize cyber-defense strategies. The agent also discovered non-intuitive policies which is encouraging to inform the design of cyber detection and defense strategies. In one instance, the agent exploited a numerical instability in the environment simulations. Overall, the training realizations appeared sensitive to the initialization of the RL agent parameters and the function mapping the action to the cumulative rewards lacked smoothness. Future work will focus on regularizing this map through the rewards function adopted. The actions chosen by the agent were simplistic enough to be practically easy to detect. In future work, an adversarial framework will be presented to the agent to promote strategies that escape existing cyber defenses.

\bibliographystyle{IEEEtran}

\begin{thebibliography}{10}
\providecommand{\url}[1]{#1}
\csname url@samestyle\endcsname
\providecommand{\newblock}{\relax}
\providecommand{\bibinfo}[2]{#2}
\providecommand{\BIBentrySTDinterwordspacing}{\spaceskip=0pt\relax}
\providecommand{\BIBentryALTinterwordstretchfactor}{4}
\providecommand{\BIBentryALTinterwordspacing}{\spaceskip=\fontdimen2\font plus
\BIBentryALTinterwordstretchfactor\fontdimen3\font minus \fontdimen4\font\relax}
\providecommand{\BIBforeignlanguage}[2]{{%
\expandafter\ifx\csname l@#1\endcsname\relax
\typeout{** WARNING: IEEEtran.bst: No hyphenation pattern has been}%
\typeout{** loaded for the language `#1'. Using the pattern for}%
\typeout{** the default language instead.}%
\else
\language=\csname l@#1\endcsname
\fi
#2}}
\providecommand{\BIBdecl}{\relax}
\BIBdecl

\bibitem{horowitz2019overview}
K.~A. Horowitz, Z.~Peterson, M.~H. Coddington, F.~Ding, B.~O. Sigrin, D.~Saleem, S.~E. Baldwin, B.~Lydic, S.~C. Stanfield, N.~Enbar \emph{et~al.}, ``An overview of distributed energy resource (der) interconnection: Current practices and emerging solutions,'' 2019.

\bibitem{olowu2020investigation}
T.~O. Olowu, S.~Dharmasena, H.~Jafari, and A.~Sarwat, ``Investigation of false data injection attacks on smart inverter settings,'' in \emph{2020 IEEE CyberPELS (CyberPELS)}.\hskip 1em plus 0.5em minus 0.4em\relax IEEE, 2020, pp. 1--6.

\bibitem{roy2019detection}
S.~D. Roy and S.~Debbarma, ``Detection and mitigation of cyber-attacks on agc systems of low inertia power grid,'' \emph{IEEE Systems Journal}, vol.~14, no.~2, pp. 2023--2031, 2019.

\bibitem{tuyen2022comprehensive}
N.~D. Tuyen, N.~S. Quan, V.~B. Linh, V.~Van~Tuyen, and G.~Fujita, ``A comprehensive review of cybersecurity in inverter-based smart power system amid the boom of renewable energy,'' \emph{IEEE Access}, vol.~10, pp. 35\,846--35\,875, 2022.

\bibitem{li2022cybersecurity}
Y.~Li and J.~Yan, ``Cybersecurity of smart inverters in the smart grid: A survey,'' \emph{IEEE Transactions on Power Electronics}, 2022.

\bibitem{sahoo2019cyber}
S.~Sahoo, T.~Dragi{\v{c}}evi{\'c}, and F.~Blaabjerg, ``Cyber security in control of grid-tied power electronic converters—challenges and vulnerabilities,'' \emph{IEEE Journal of Emerging and Selected Topics in Power Electronics}, vol.~9, no.~5, pp. 5326--5340, 2019.

\bibitem{liu2015analyzing}
R.~Liu, C.~Vellaithurai, S.~S. Biswas, T.~T. Gamage, and A.~K. Srivastava, ``Analyzing the cyber-physical impact of cyber events on the power grid,'' \emph{IEEE Transactions on Smart Grid}, vol.~6, no.~5, pp. 2444--2453, 2015.

\bibitem{zhang_2021}
Z.~{Zhang}, Y.~{Mishra}, D.~{Yue}, C.~{Dou}, B.~{Zhang}, and Y.~C. {Tian}, ``Delay-tolerant predictive power compensation control for photovoltaic voltage regulation,'' \emph{IEEE Transactions on Industrial Informatics}, vol.~17, no.~7, pp. 4545--4554, 2021.

\bibitem{ossenbuhl2015towards}
S.~Ossenb{\"u}hl, J.~Steinberger, and H.~Baier, ``Towards automated incident handling: How to select an appropriate response against a network-based attack?'' in \emph{2015 Ninth International Conference on IT Security Incident Management \& IT Forensics}.\hskip 1em plus 0.5em minus 0.4em\relax IEEE, 2015, pp. 51--67.

\bibitem{gawlowicz2019ns}
P.~Gaw{\l}owicz and A.~Zubow, ``Ns-3 meets openai gym: The playground for machine learning in networking research,'' in \emph{Proceedings of the 22nd International ACM Conference on Modeling, Analysis and Simulation of Wireless and Mobile Systems}, 2019, pp. 113--120.

\bibitem{single_area_agc}
J.~Li and T.~Yu, ``Deep reinforcement learning based multi-objective integrated automatic generation control for multiple continuous power disturbances,'' \emph{IEEE Access}, vol.~8, pp. 156\,839--156\,850, 2020.

\bibitem{multi_area_agc}
Z.~Yan and Y.~Xu, ``A multi-agent deep reinforcement learning method for cooperative load frequency control of a multi-area power system,'' \emph{IEEE Transactions on Power Systems}, vol.~35, no.~6, pp. 4599--4608, 2020.

\bibitem{kundur2022power}
P.~S. Kundur and O.~P. Malik, \emph{Power system stability and control}.\hskip 1em plus 0.5em minus 0.4em\relax McGraw-Hill Education, 2022.

\bibitem{cui2022reinforcement}
W.~Cui, Y.~Jiang, and B.~Zhang, ``Reinforcement learning for optimal primary frequency control: A lyapunov approach,'' \emph{IEEE Transactions on Power Systems}, vol.~38, no.~2, pp. 1676--1688, 2022.

\bibitem{nguyen2020electric}
T.~Nguyen, S.~Wang, M.~Alhazmi, M.~Nazemi, A.~Estebsari, and P.~Dehghanian, ``Electric power grid resilience to cyber adversaries: State of the art,'' \emph{IEEE Access}, vol.~8, pp. 87\,592--87\,608, 2020.

\bibitem{sutton_barto_2018}
R.~S. Sutton and A.~Barto, \emph{\BIBforeignlanguage{en}{Reinforcement learning: an introduction}}, second edition~ed., ser. Adaptive computation and machine learning.\hskip 1em plus 0.5em minus 0.4em\relax Cambridge, Massachusetts London, England: The MIT Press, 2018.

\bibitem{yu2022surprising}
C.~Yu, A.~Velu, E.~Vinitsky, J.~Gao, Y.~Wang, A.~Bayen, and Y.~Wu, ``The surprising effectiveness of ppo in cooperative multi-agent games,'' \emph{Advances in Neural Information Processing Systems}, vol.~35, pp. 24\,611--24\,624, 2022.

\bibitem{schulman2017proximal}
J.~Schulman, F.~Wolski, P.~Dhariwal, A.~Radford, and O.~Klimov, ``{Proximal Policy Optimization Algorithms},'' 2017.

\bibitem{mnih2016asynchronous}
V.~Mnih, A.~P. Badia, M.~Mirza, A.~Graves, T.~Lillicrap, T.~Harley, D.~Silver, and K.~Kavukcuoglu, ``Asynchronous methods for deep reinforcement learning,'' in \emph{International conference on machine learning}.\hskip 1em plus 0.5em minus 0.4em\relax PMLR, 2016, pp. 1928--1937.

\bibitem{stable-baselines3}
\BIBentryALTinterwordspacing
A.~Raffin, A.~Hill, A.~Gleave, A.~Kanervisto, M.~Ernestus, and N.~Dormann, ``Stable-baselines3: Reliable reinforcement learning implementations,'' \emph{Journal of Machine Learning Research}, vol.~22, no. 268, pp. 1--8, 2021. [Online]. Available: \url{http://jmlr.org/papers/v22/20-1364.html}
\BIBentrySTDinterwordspacing

\end{thebibliography}

\end{document}